\documentclass[letterpaper, 10 pt, journal, twoside]{IEEEtran}
\IEEEoverridecommandlockouts

\usepackage{times}
\usepackage{soul}
\usepackage{url}
\usepackage[hidelinks]{hyperref}
\usepackage{graphicx}
\usepackage{amsmath}
\usepackage{amssymb}
\usepackage{booktabs}
\usepackage{multirow}
\usepackage{algorithm}
\usepackage{algpseudocode}

\usepackage{multirow}
\usepackage{array}
\usepackage{xcolor}
\usepackage{cite}
\usepackage{float}

\algrenewcommand\algorithmicindent{1.0em}%

\begin{document}

\author{Yan Ding, Xiaohan Zhang, Xingyue Zhan, and Shiqi Zhang%
\thanks{Manuscript received: September 9, 2021; Revised December 29, 2021; Accepted January 30, 2022.}
\thanks{This paper was recommended for publication by Editor Hanna Kurniawati upon evaluation of the Associate Editor and Reviewers' comments.
This work has taken place at the Autonomous Intelligent Robotics (AIR)
Group, SUNY Binghamton. AIR research is supported in part by grants
from the NSF (NRI-1925044), Ford, OPPO, and SUNY RF.} 
\thanks{The authors are with the Department of Computer Science, SUNY Binghamton, Binghamton NY 13902
        {\tt\footnotesize yding25@binghamton.edu}}%
\thanks{Digital Object Identifier (DOI): see top of this page.}
}

\title{Learning to Ground Objects for Robot Task and Motion Planning
}

\maketitle

\begin{abstract}
Task and motion planning (TAMP) algorithms have been developed to help robots plan behaviors in discrete and continuous spaces. 
Robots face complex real-world scenarios, where it is hardly possible to model all objects or their physical properties for robot planning (e.g., in kitchens or shopping centers). 
In this paper, we define a new object-centric TAMP problem, where the TAMP robot does not know object properties (e.g., size and weight of blocks). 
We then introduce Task-Motion Object-Centric planning ({\bf TMOC}), a grounded TAMP algorithm that learns to ground objects and their physical properties with a physics engine. 
TMOC is particularly useful for those tasks that involve dynamic complex robot-multi-object interactions that can hardly be modeled beforehand. 
We have demonstrated and evaluated TMOC in simulation and using a real robot. 
Results show that TMOC outperforms competitive baselines from the literature in cumulative utility. 
\end{abstract}


\section{Introduction}
\label{sec:introduction}
\IEEEPARstart{R}{obots} that operate in the real world need to plan at both task and motion levels. .
At the task level, the robot computes a sequence of symbolic actions in a discrete space to achieve long-term goals~\cite{ghallab2004automated}. 
At the motion level, the symbolic actions are implemented in a continuous space, and the computed trajectories can be directly applied to the real world~\cite{choset2005principles}.
Planning at task and motion levels at the same time is challenging ~\cite{kaelbling2013integrated,garrett2018sampling,toussaint2015logic,dantam2018incremental,chitnis2016guided,lo2020petlon}, resulting in the so-called integrated \emph{task and motion planning} (\textbf{TAMP}) problem~\cite{garrett2021integrated,lagriffoul2018platform}, where the main challenge is to achieve task-level goals while maintaining motion-level feasibility. 

\begin{figure}[t]
\vspace{.7em}
\centering
\includegraphics[width=1.0\columnwidth]{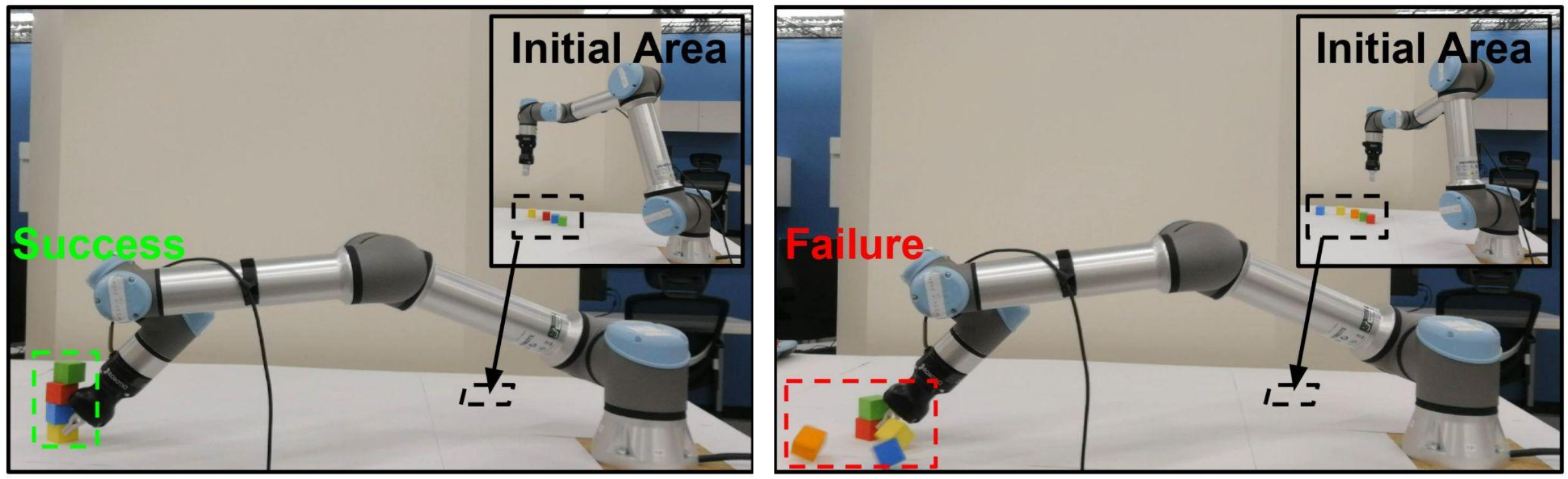}
\vspace{-1.5em} 
\caption{A robot performing ``stack-and-push'' tasks by repeatedly building a stack of blocks and pushing the stack to a goal area. 
There is a trade-off between stability and efficiency (higher stacks are less stable) that the robot must learn from trial and error. 
It is a TAMP problem, e.g., the stacking behaviors require both task planning and motion planning.}
\label{fig:demo_introduction}
\end{figure}

Planning algorithms (at the task, motion, or both levels) frequently assume that a world model is provided beforehand, including how the world reacts to robot behaviors (i.e., world dynamics). 
However, many real-world scenarios are very complex, making it hardly possible to model all objects or their physical properties at planning time, e.g.,~kitchens and shopping centers. 
In such scenarios, modeling how a robot interacts with multiple objects is even more challenging, e.g., holding a stack of plates, cutting onions, and squeezing through a crowd. 
Fig.~\ref{fig:demo_introduction} shows a ``\textbf{stack-and-push}'' task as an example scenario with complex multi-object interactions, where a robot repeatedly builds a block tower and then pushes it to a goal area. 
Aiming to move all blocks to the goal area, the robot needs to learn object properties (e.g., size and weight), how to grasp the blocks (which depends on the object properties), and a stack-and-push strategy (e.g., how many blocks to be stacked together). 

The \textbf{first contribution} of this work is a new grounded, object-centric TAMP framework, called task and motion Planning with Physics-based Simulation (\textbf{PPS}). 
The uniqueness of PPS lies in the inclusion of a physics engine for grounding objects and their physical properties. 
Object grounding enables the robot to collect simulated interaction experiences for policy learning purposes. 
The PPS framework is general enough to accommodate different TAMP problems that vary in model completeness.

Our \textbf{second contribution} is an algorithm, called {Task-Motion Object-Centric} planning~(\textbf{TMOC}), that addresses a challenging PPS problem where \emph{object properties are not provided}.
TMOC is particularly useful for those TAMP domains that involve complex multi-object dynamic interactions. 
Our TMOC robot can ground objects in a high-fidelity physics engine, learn object properties to facilitate the grounding, and improve its task-motion planning skills. 

We have demonstrated and evaluated TMOC in simulation (where the robot uses a container to move objects), and using a real robot (where the robot conducts stack-and-push tasks).
The simulation and real-world scenarios share the challenge of robot task-motion planning with unknown object properties. 
From the results, we see that TMOC outperforms a set of baselines from literature in cumulative utility.
Finally, we demonstrate the learning process of a real robot that uses TMOC for task-motion planning.

\section{Related Work}

There is a long history of developing planning algorithms in robotics research. 
We summarize three research areas that are the most relevant to this research, namely TAMP, symbol learning for robot planning, and sim-to-real transfer. 

\subsection{Task and Motion Planning (TAMP)}
Broadly, any robots that plan behaviors at a high level and operate in the real world would need algorithms for TAMP.
However, it is not until recently that TAMP has been used as a term to refer to the algorithms that interleave the processes of task planning and motion planning~\cite{garrett2021integrated,lagriffoul2018platform,lo2020petlon}. 
Early examples include aSyMov~\cite{cambon2009hybrid} and Semantic Attachment~\cite{dornhege2009semantic}. 
Hierarchical planning in the now (HPN)~\cite{kaelbling2010hierarchical} has been extended to model the uncertainty in action outcomes and observability~\cite{kaelbling2013integrated}. 
FFRob directly conducts task planning over a set of samples generated in the configuration space~\cite{garrett2018ffrob}.
Probabilistically complete TAMP was achieved using constraint satisfaction methods~\cite{dantam2018incremental}. 
Wang et al. used a policy synthesis approach to account for uncontrollable agents (such as humans)~\cite{wang2016task}. 
Off-the-shelf task-motion planners can be integrated using a planner-independent interface~\cite{srivastava2014combined}. 
Optimization methods have been applied to TAMP domains, where the goal is specified with a cost function~\cite{toussaint2015logic}.
In comparison to planning methods (including TAMP) that assume knowledge of complete world models, this work considers robots that compute world models (e.g., objects' physical properties) through perception, and actively estimate the current world state, which renders ``grounding'' necessary.

Very recently, researchers developed a visually grounded TAMP approach, called GROP, for mobile manipulation~\cite{zhang2022visually}. 
Compared with GROP that uses computer vision techniques to learn a state mapping function, TMOC (ours) learns to ground objects and their physical properties for TAMP tasks. 

\subsection{Symbol Learning for Robot Planning}
Researchers have developed symbol learning methods for robot planning. 
Konidaris et al. (2018) focused on learning symbols to construct representations that are provably capable of evaluating plans composed of sequences of those actions~\cite{konidaris2018skills}, assuming the robot was equipped with a collection of high-level actions.
Their mobile manipulator learned a grounded symbol that indicates that a cupboard door is open, which is necessary for determining whether or not the robot can pick up a bottle from the cupboard. 
Gopalan et al. (2020) introduced natural language into the loop, and showed that
the learned symbols can enable a robot to learn to map natural language to goal-based planning with only trajectories as supervision~\cite{gopalan2020simultaneously}.
The above-mentioned methods assumed that low-level motion primitives are provided (as skills or demonstrations), and the robot needs to learn a symbolic representation that can be used for planning to accomplish different complex tasks. 
In this work, we assume those symbols are provided, but there are no models about the symbols' physical meanings in the real world.

\subsection{Sim-to-Real Transfer} This work requires a high-fidelity simulator, which is related to ``sim-to-real'' methods to enable agents to learn from simulation for operating in the real world~\cite{jakobi1995noise}.
Focusing on addressing the reality gap, there are at least two families of sim-to-real methods.
One intentionally adds different forms of noise into the simulation environment to learn policies that are robust enough to work in the real world~\cite{tan2018sim,peng2018sim,molchanov2019sim,yu2017preparing,pinto2017robust}; the other actively updates parameters of the simulator toward generating realistic experience for policy learning purposes~\cite{farchy2013humanoid,chebotar2019closing,zhu2017fast,jeong2020self}.
This work falls into the second category of methods from the perspective of updating the simulator's parameters using real-world data.
The main difference is that our method is placed in the TAMP context.
As a result, our proposed method faces two simultaneous, interdependent reality gaps at task and motion levels, respectively.

This paper enables robots to plan at both task and motion levels in object-centric domains where the object properties can be unknown.
Our developed approach is particularly suitable for task-motion domains that involve complex multi-object dynamic interactions.
Next, we define the grounded TAMP problem and then describe our algorithm.

\section{Framework and Problem Statements}
\label{label:definitions}

In this section, we first define an object-centric TAMP framework, and then describe how the framework accommodates a few TAMP problems.

\begin{figure*}[t]
\centering
\vspace{.5em}
\includegraphics[width=1.75\columnwidth]{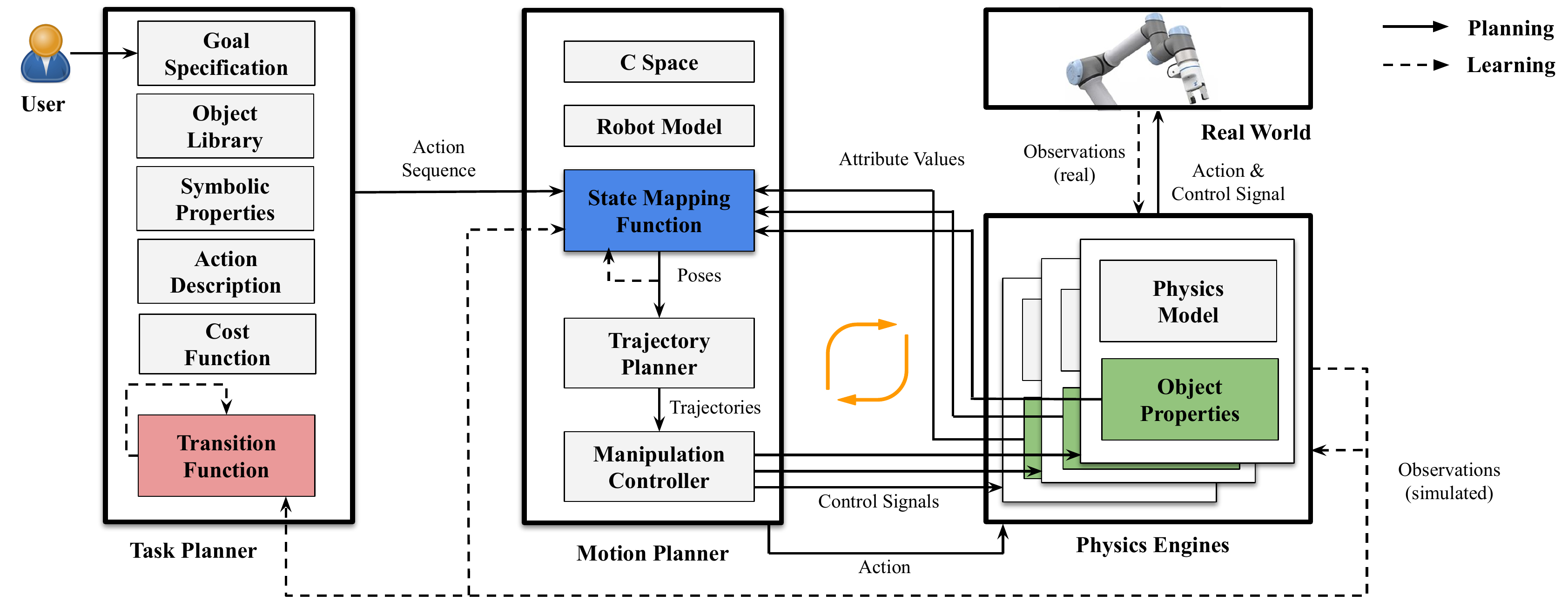}
\caption{
An overview of the TMOC algorithm for our PPS problem with unknown object properties ($L$ in green), state mapping function ($Y$ in blue), and transition function ($T$ in red). 
The main components of TMOC include a task planner for sequencing symbolic actions, a motion planner for computing motion trajectories, and a physics engine for simulating multiple grounded worlds.
TMOC takes object library, task domain description, goal specification, cost function, motion domain description and state mapping function as input.
TMOC aims to compute task-motion plans to achieve task-level goals while maximizing cumulative utilities.
Using TMOC, a robot learns object properties ($L$) from task-completion experience, and leverages the learned properties to further improve its task-motion planning skills ($Y$ and $T$) over time.
} 
\label{fig:framework}
\end{figure*}

\subsection{Framework}\label{sec:framework}

\emph{Planning with Physics-based Simulation} (\textbf{PPS}) is a task-motion planning framework\footnote{Researchers have developed TAMP frameworks, such as PETLON for navigation domains~\cite{lo2020petlon}, and FFRob~\cite{garrett2018ffrob} for heuristics-based TAMP. 
The uniqueness of PPS, as an object-centric TAMP framework, is attributed to its inclusions of a state mapping function, and a high-fidelity simulator.}, and is defined as tuple 
$$
    \langle Obj, D^t, D^m, Y\rangle, 
$$
where $Obj$ is a finite set of objects, and the other components are described next. 

\vspace{.5em}
\noindent
\textbf{Task-level Domain Description}
$D^t:\langle E,A,T\rangle$, where $E$ is a finite set of symbolic (discrete) properties, and $E^{obj} \subseteq E$ includes symbolic properties that are applicable to $obj\in Obj$. 
Each property $e\in E^{obj}$ has a finite domain of values, referred to as $V^{obj}_e$. 
For instance, in a manipulation domain, object $box\in Obj$ has a property of $loaded\in E^{box}$ that has domain $V^{box}_{loaded}=\{true, false\}$. 
A state $s$ is an assignment of each object's applicable symbolic properties. 
$S$ denotes the set of possible states in symbolic forms. 
We use $s^{init}\in S$ and $S^G \subseteq S$ to represent the initial state and the set of goal states, respectively.

$A$ is a finite set of object-centric actions, and action $a \in A$ is defined by its preconditions and effects. 
State transition model $T(s,a,s')$ defines how an action leads (deterministic or probabilistic) transitions.
When action costs are considered in task planning, a task planner is able to select a plan out of all the plans that satisfy the goal conditions toward minimizing the overall cost. 
In this paper, at the task level, we are concerned with object-centric, cost-sensitive, goal-conditioned planning under uncertainty. 

\vspace{.5em}
\noindent
\textbf{Motion-level Domain Description}
$D^m: \langle L,C \rangle$, where $L$ is a finite set of continuous properties. 
$L$ is the counterpart of $E$ at the task level.
$L^{obj} \subseteq L$ includes properties that are applicable to $obj\in Obj$, and properties $l \in L^{obj}$ is a real number. 
Example continuous properties include an object's size, position, and weight. 
The configuration space $C$ is the set of all possible configurations. 
We use $\xi$ to represent a motion trajectory, where the trajectory connects the robot's current configuration ($x^{init}$) to any configuration in the target space ($x^{goal} \in C$). 

\vspace{.5em}
\noindent
\textbf{State Mapping Function} $Y$ maps a task-level state to a motion-level pose: $x \leftarrow Y(s)$, where $s$ is a task-level state and $x\in C$ is a configuration.
$Y^{-1}$ is the inverse function of $Y$, and outputs a symbolic state $s$ given a motion-level configuration. 
Here we assume $Y^{-1}$ can be derived from $Y$, and is thus omitted from the definition of PPS. 

PPS is a general-purpose TAMP framework that is able to accommodate different task-motion planning problems. 
A realization of a PPS framework requires a task planning system, a motion planning system, and a physics-based simulation (physics engine) system.
One can leverage the physics engine of PPS to ground different components of the real world. 

Next, we define a challenging PPS problem, and discuss how this problem connects to a few existing TAMP problems.

\subsection{Problem Statements}
\label{sec:problems}

In this paper, we aim to address the PPS problem with the following set of functions being unknown: 
$$
    \{L, Y, T\}
$$

Consider a ``\textbf{stack-and-push}'' domain: A robot does not know block size and weight (about $L$), where to place the gripper for grasping (about $Y$), and how stable a stack of $N$ blocks is (about $T$). 
This is a challenging problem, because planning at task and motion levels highly depends on the estimation of object properties, and the task-level strategy further depends on the robot's motion-level performance. 
We use the stack-and-push domain in real-robot experiments, where the robot aims to move blocks from an initial area to a goal area as quickly and stably as possible.

The PPS framework is general enough to accommodate different TAMP problems (in addition to the above-mentioned problem that is the focus of this paper). 
For instance, when all components of PPS are known, it corresponds to a standard TAMP problem, e.g.,~\cite{lo2020petlon,ding2020task}. 
When only transition function $T$ is unknown, it corresponds to existing research, such as~\cite{wang2021learning, dantam2018incremental}. 
Planning domains with unknown physical properties $L$~\cite{jiang2019task}, and planning domains with unknown state mapping function $Y$~\cite{liang2020learning} can be modeled as PPS problems as well. 

The input of a PPS algorithm includes a PPS domain in the form of $\langle Obj, D^t, D^m, Y\rangle$, and a set of task-level goal states $S^G$.
Our utility function incorporates action costs, success bonus, and failure penalty.
A PPS algorithm aims to compute task-motion plans to achieve task-level goals while maximizing cumulative utilities. 

Next, we focus on our PPS problem with unknown object properties $L$ (where the robot is motivated to learn $Y$ and $T$ accordingly), and develop algorithm TMOC to enable robots to learn to plan at task and motion levels. 

\section{Algorithm}\label{sec:algorithm}
In this section, we present the main contribution of this research, \textbf{TMOC} (short for ``task-motion object-centric'' planning), a grounded task-motion planning algorithm for addressing the PPS problem described in Section~\ref{sec:problems}, where object properties ($L$), state mapping function ($Y$), and transition function ($T$) are unknown.
TMOC includes three main components of a task planner, a motion planner, and a physics engine, as illustrated in Fig.~\ref{fig:framework}. 

Algorithm~\ref{alg:TMOC} presents the control loops of TMOC. 
The input of TMOC includes a PPS domain in the form of $\langle Obj, D^t, D^m, Y\rangle$, and a set of task-level goal states $S^G$.
TMOC learns $L$, $Y$, and $T$ in each iteration, where the robot completes a task \emph{once} in the real world and \emph{$N$ times} in simulation.
In each iteration, each of $L$, $Y$, and $T$ is learned under the current estimation of the other two, while a TMOC agent plans at task and motion levels.

\subsection{TMOC algorithm}

A realization of a PPS framework requires task planning system $P^t$, motion planning system $P^m$, and physics-based simulation system $\mathcal{M}$. 
$P^t$ takes task domain $D^t$ together with the current state $s$ (which can be derived from $D^t$) as input and generates an action sequence. 
Specifically, $P^m$ takes the initial configuration $x^{init}$, goal configuration set $X^G$, and $D^t$ as input, and generates motion trajectory $\xi$. 
$\mathcal{M}$ takes $\xi$, object set $Obj$, and their physics-relevant properties $L$ as input, and generates $L'$ denoting the resulting properties of $Obj$. 
From $L'$, one can infer motion planning domain $D^m$, which can be further used for computing symbolic state $s$ using the function $Y^{-1}$. 

\indent
\vspace{-1em}
\begin{algorithm}[h]
\small
\caption{TMOC algorithm}\label{alg:TMOC}
\textbf{Require:} {$P^t$, $P^m$, and $\mathcal{M}$}\\
\textbf{Input:} {$\langle Obj, D^t, D^m, Y\rangle$, $S^G$ and $Cost$}
\begin{algorithmic}[1]
\State {Initialize $T$ optimistically}\label{alg:init_T}
\State {Initialize $N$ simulated worlds and each one has $L_i$ (particle) with a uniform weight $w_i$, where $i=0, 1,\cdots, N-1$}\label{alg:init_N}
\State {Initialize an empty experience pool $Pool$}\label{alg:init_Pool}
\While{True}\Comment{start working on a new trial}
\State {$p$ $\leftarrow$ $P^t(\mathcal{D}^t, s^{init}, S^G, Cost)$, where $D^t$ includes $T$}\label{alg:compute_tp}\Comment{this is task planning}
\For{each action $\langle s, a\rangle$ in $p$}\label{alg:exec_tp_s}
\For{$i=0, 1,\cdots, N\!-\!1$}\label{alg:real_s}
\State {Compute $Y_i$ using $L_i$ for its simulated world}\label{alg:update_Y}
\State {Generate a feasible pose $x_i$ for $\langle s, a\rangle$ using $Y_i$}\label{alg:pose_sim}
\EndFor
\State {Compute a pose $x$ for the real world by weighted \indent averaging poses $\{x_0, x_1, \cdots, x_{N-1}\}$}\label{alg:pose_real}
\State {$\xi \leftarrow P^m(x_{curr}, x, D^m)$}\label{alg:traj_real}\Comment{this is motion planning}
\State {Execute $\xi$ in the real world and then obtain a resulting \indent state $s'$}\label{alg:compute_weight_real}
\State {$Pool=Pool$ $\cup$ $\{\langle s, a, s' \rangle\}$}\label{alg:update_Pool_real}\label{alg:real_e}
\For{$i=0, 1,\cdots, N\!-\!1$}\label{alg:sim_s}
\State {$\xi_i \leftarrow P^m(x_{curr}, x_i, D^m$)}\label{alg:traj_sim}
\State {$L_i' \leftarrow \mathcal{M}(\xi_i, Obj, L_i)$, where $L_i$ is in $D^m$}\label{alg:new_L}
\State {Infer a resulting state $s_i'$ using $L_i'$ and $Y_i^{-1}$, and then \indent \hspace{-1.5em} \indent $L_i \leftarrow L_i'$}\label{alg:new_state_sim}
\State {$Pool=Pool$ $\cup$ $\{\langle s, a, s_i' \rangle\}$}\label{alg:update_Pool_sim}
\State {Compute weight $w_i$ of $L_i$ using $s'$ and $s_i'$
}\label{alg:compute_weight}
\EndFor\label{alg:sim_e}
\State {Update $L_i$ from the discrete distribution given by \indent $\{w_0, w_1, \cdots, w_{N-1}\}$}\label{alg:update_L}
\State {Update $T$ using $Pool$}\label{alg:update_T}
\EndFor\label{alg:exec_tp_e}
\EndWhile
\end{algorithmic}
\end{algorithm}

\vspace{.5em}
\noindent
\textbf{Structure of Algorithm TMOC: }
{Lines \ref{alg:init_T}$\sim$\ref{alg:init_Pool}} are for initializing data structures. 
{Lines \ref{alg:compute_tp}$\sim$\ref{alg:exec_tp_e}} form a complete iteration. 
In each iteration, Line~\ref{alg:compute_tp} computes a task-level plan, and the functions of $Y$, $L$, and $T$ are updated from trial and error in Lines~\ref{alg:update_Y}, \ref{alg:update_L}, and \ref{alg:update_T}, respectively. 
{Lines \ref{alg:real_s}$\sim$\ref{alg:real_e}} are for executing an action in the real world; and Lines \ref{alg:sim_s}$\sim$\ref{alg:sim_e} are for executing the same action in simulation. 
TMOC is a life-long learning algorithm, and does not have termination conditions. Next, we look into individual lines of TMOC. 

Line~\ref{alg:init_T} initializes transition function $T$ ``optimistically''. 
A transition function ($T$) indicates how reliable an action is. Initializing $T$ optimistically means that the robot believes it always gets the desired results after performing an action, e.g., manipulation and push actions are always successful.
This initialization strategy is inspired by the R-max algorithm~\cite{brafman2002r}, and encourages exploration in the early learning phase. 
TMOC initializes $N$ simulated worlds, where each is specified by a set of physics-relevant properties $L_i$, e.g., length, width, and density of blocks~{(Line~\ref{alg:init_N})}.
Each simulated world is associated with a weight $w_i$, and the weights are uniformly initialized. 
An empty experience pool $Pool$ is initialized to store observations (i.e., actions and their resulting states) after action executions~{(Line~\ref{alg:init_Pool})}.

{Line~\ref{alg:compute_tp}} computes an optimal action sequence $p$ using the task planning system $P^t$ given the utility function, which takes into account transition function $T$ (e.g., how reliable grasps are) and cost function $Cost$ (e.g., how long it takes a robot to grasp an object).

{Lines \ref{alg:real_s}$\sim$\ref{alg:real_e}} describe how action $\langle s, a\rangle$ is implemented in the \textbf{real world}.
From $L_i$, we can get the configuration space, based on which $Y_i$ can be computed using motion planning methods~({Line~\ref{alg:update_Y}}). 
Thus, a feasible pose $x_i$ for action $\langle s, a\rangle$ can be generated using $Y_i$~({Line~\ref{alg:pose_sim}}). 
A new pose $x$ for the real world is  computed by weighted averaging poses $x_i$, where $i=0, 1, \cdots, N-1$~({Line~\ref{alg:pose_real}}). 
A trajectory $\xi$ is computed using the motion planner $P^m$, which takes $x_{curr}$, computed $x$, and $D^m$ as input, where $x_{curr}$ refers to the current pose of the robot~({Line~\ref{alg:traj_real}}).
$\xi$ is executed in the real world, and thus a resulting state $s'$ of action $\langle s, a\rangle$ is obtained~({Line~\ref{alg:compute_weight_real}}).
Finally, tuple $\langle s, a, s'\rangle$ is added to experience pool $Pool$~({Line~\ref{alg:update_Pool_real}}).

{Lines \ref{alg:sim_s}$\sim$\ref{alg:sim_e}} explain how action $\langle s, a\rangle$ is implemented in all \textbf{simulated worlds} (one iteration for each world).
A trajectory $\xi_i$ is computed using motion planning system $P^m$~({Line~\ref{alg:traj_sim}}). 
Then, a set of resulting properties $L'_i$ is obtained using physical-based simulation system $\mathcal{M}$~({Line~\ref{alg:new_L}}).
Thus, a resulting state $s_i'$ can be inferred using $L_i'$ and the reverse function of $Y_i$, and $L_i$ is also updated by $L_i'$~({Line~\ref{alg:new_state_sim}}).
Finally, tuple $\langle s, a, s_i' \rangle$ is added to $Pool$~({Line~\ref{alg:update_Pool_sim}}).

{Lines \ref{alg:compute_weight}}, {\ref{alg:update_L} $\sim$ \ref{alg:update_T}} update $L$ and $T$ using action completion experience.
Specifically, each weight $w_i$ of $L_i$ can be computed based on the resulting state from the real and simulated worlds~({Line~\ref{alg:compute_weight}}).
Similar to the particle filter~\cite{gordon1993novel}, a new $L_i$ is sampled from the discrete distribution given by $w_i$, where $i = 0, 1, \cdots, N-1$~({Line~\ref{alg:update_L}}).
Besides, $T$ is updated using the experience pool $Pool$~({Line~\ref{alg:update_T}}).


TMOC enables robots to plan at both task and motion levels in object-centric domains where the objects' physical properties ($L$) are unknown (and thus $T$ and $Y$ are unknown).
TMOC is particularly useful for those tasks that involve dynamic complex robot-multi-object interactions that can hardly be modeled beforehand. 
Next, we focus on the evaluation of TMOC in both simulation and the real world.

\begin{figure*}[t]
\centering
\vspace{0.5em}
\includegraphics[width=2.0\columnwidth]{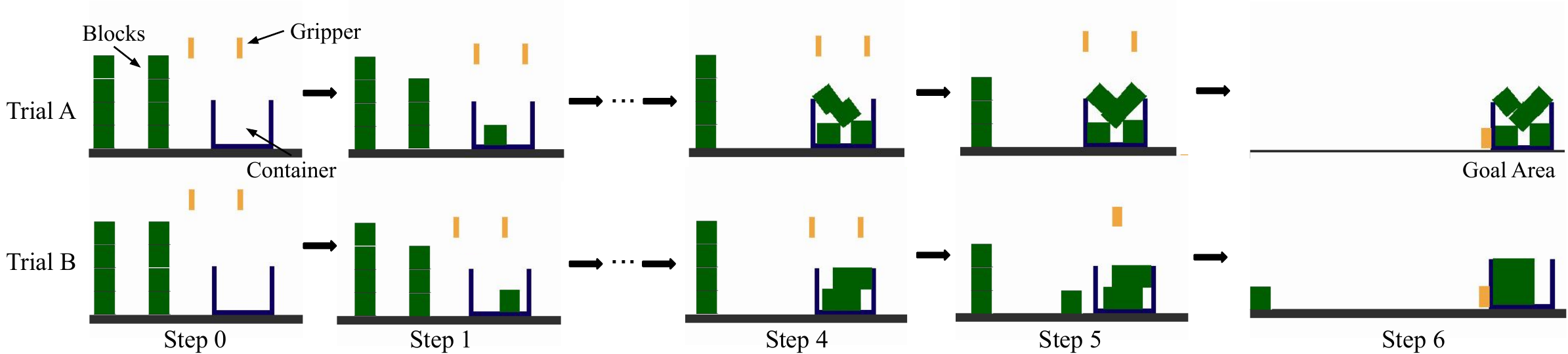}
\caption{
A robot needs to stack a set of blocks (Steps $0\sim6$ of Trial A and B) and then push them to the goal area with a container. 
The more blocks in the container, the less stable it is, resulting in a trade-off between work efficiency and success rate.
At the task level, the robot is concerned about how many blocks (of initially unknown sizes and weights) can be placed in the container. 
Step 4 of trials A and B demonstrate two such situations, where the robot needs to decide if another block can be added to the container. 
Making this decision is difficult due to the complex multi-object interactions.
The two trials give different results as shown in Step 5 of trials A and B.
}
\label{fig:demo_pybox2D}
\end{figure*}

\subsection{Algorithm Instantiation}\label{sec:instantiation}
\noindent\textbf{Task Planner:}
Our task planner $P^t$ is implemented using Answer Set Programming (ASP), which is a popular declarative language for knowledge representation and reasoning.
ASP has been applied to task planning~\cite{lifschitz2002answer,amiri2019augmenting,lo2020petlon,jiang2019task}. 
In out domain, predicate \texttt{is\_holding(R1,B1)} is used to specify block \texttt{B1} being in the robot hand \texttt{R1}.
We model three manipulation actions, including \texttt{pickup}, \texttt{stack}, and \texttt{push}. 
For instance, action \texttt{pickup} is used to help the robot arm pick up the target block from an initial location, where constraints, such as ``\texttt{stack} is allowed only if a target block is in the robot hand'', have been modeled as well. 
An example goal specification can be ``all blocks are in the container, and containers are at the goal location''.
Table~\ref{table:action} defines three actions by their preconditions and effects, where $R$, $B$, $C$, and $L$ stand for robot, block, container, and location, respectively.

\begin{table}
\footnotesize
\centering
\caption{Action knowledge in our system, organized into preconditions and effects.}\label{table:action}
\begin{tabular}{ cccc } 
\toprule
\textbf{Action} & \textbf{Precondition} & \textbf{Effect} \\
\midrule
\multirow{2}{8.0em}{$pickup(R,B,L)$} & \multirow{1}{7.1em}{$is\_in(B,L)$} & $is\_holding(R,B)$ \\ 
                                   & \multirow{1}{7.1em}{$hand\_empty(R)$} & $\neg hand\_empty(R)$ \\
\hline
\multirow{3}{8.0em}{$stack(R,B,C)$} & \multirow{3}{7.1em}{$is\_holding(R,B)$} & $\neg is\_holding(R,B)$\\ 
                                   &  & $hand\_empty(R)$ \\
                                   &  & $at\_container(B,C)$ \\ 
\hline
\multirow{2}{8.0em}{$push(R,C,L1,L2)$} & \multirow{1}{7.1em}{$hand\_empty(R)$} & $hand\_empty(R)$ \\ 
                                   & \multirow{1}{7.1em}{$is\_in(C,L1)$} & $is\_in(C,L2)$ \\ 
\bottomrule
\end{tabular}
\vspace{-1em}
\end{table}

\vspace{.8em}

\noindent\textbf{Motion Planner:}
At the motion level, given a configuration space, a roadmap is firstly built for the robot. 
A trajectory planner then generates a desired continuous and collision-free trajectory with minimal traveling distance using optimization algorithms, RRT* in our case~\cite{karaman2011sampling}. 
The trajectory includes a set of poses, and the trajectory is delivered to the manipulation controller, along with the robot's current pose.

\section{Experiments} \label{sec:exp}
We have conducted experiments both in simulation and using a real robot. 
In simulation experiments, we focus on statistically comparing the performances of TMOC and a set of competitive baselines using results collected from large numbers of trials. 
The comparisons are based on cumulative utility, which is a combined measurement that incorporates action costs, success bonus, and failure penalty.\footnote{The action cost of $pickup$, $stack$, and $push$ are 15, 10 and 30, respectively, where a cost technically corresponds to a negative reward. The bonus of successfully stacking a block, and pushing a container to a goal location is 40 and 80, respectively. The failure penalty is 50.}
In real-world experiments, we illustrate a complete learning process of a real robot arm performing ``stack and push'' tasks. 

\subsection{Experiments in Simulation}

We used an open-source 2D physics-based simulator called Pybox2D~\cite{hunt2019introduction} to evaluate the performance of TMOC.
We decided to use a 2D simulator instead of a 3D one because 3D simulators (e.g., Gazebo~\cite{koenig2004design} and PyBullet~\cite{coumans2016pybullet}) are generally computationally less efficient. 
Our agent needs to learn to interact with objects (with unknown properties) at both task and motion levels, which requires considerable interaction experience, so we selected a 2D simulator that allows running experiments extensively.

Fig.~\ref{fig:demo_pybox2D} illustrates how a robot gripper uses a container to move eight blocks to a goal area\footnote{In this paper, we assume blocks and the container are not too heavy to be manipulated by the robot. Besides, the blocks are small enough to be graspable by the robot arm.}. 
The robot knows that all blocks share the same density, but must estimate their sizes for task-motion planning.
The robot might fail in loading and unloading a block, and in blocks falling off from the container. 
Here we assume a ``helper'' helps move blocks out of the container in the goal area before the gripper and container are moved back. 
We use this task to capture complex contact-based interactions among multiple objects, whereas the interactions in ``stack-and-push'' scenarios (used for real-robot experiments) do not go beyond two objects. 

\begin{figure*}[t]
\centering
\includegraphics[width=.69\columnwidth]{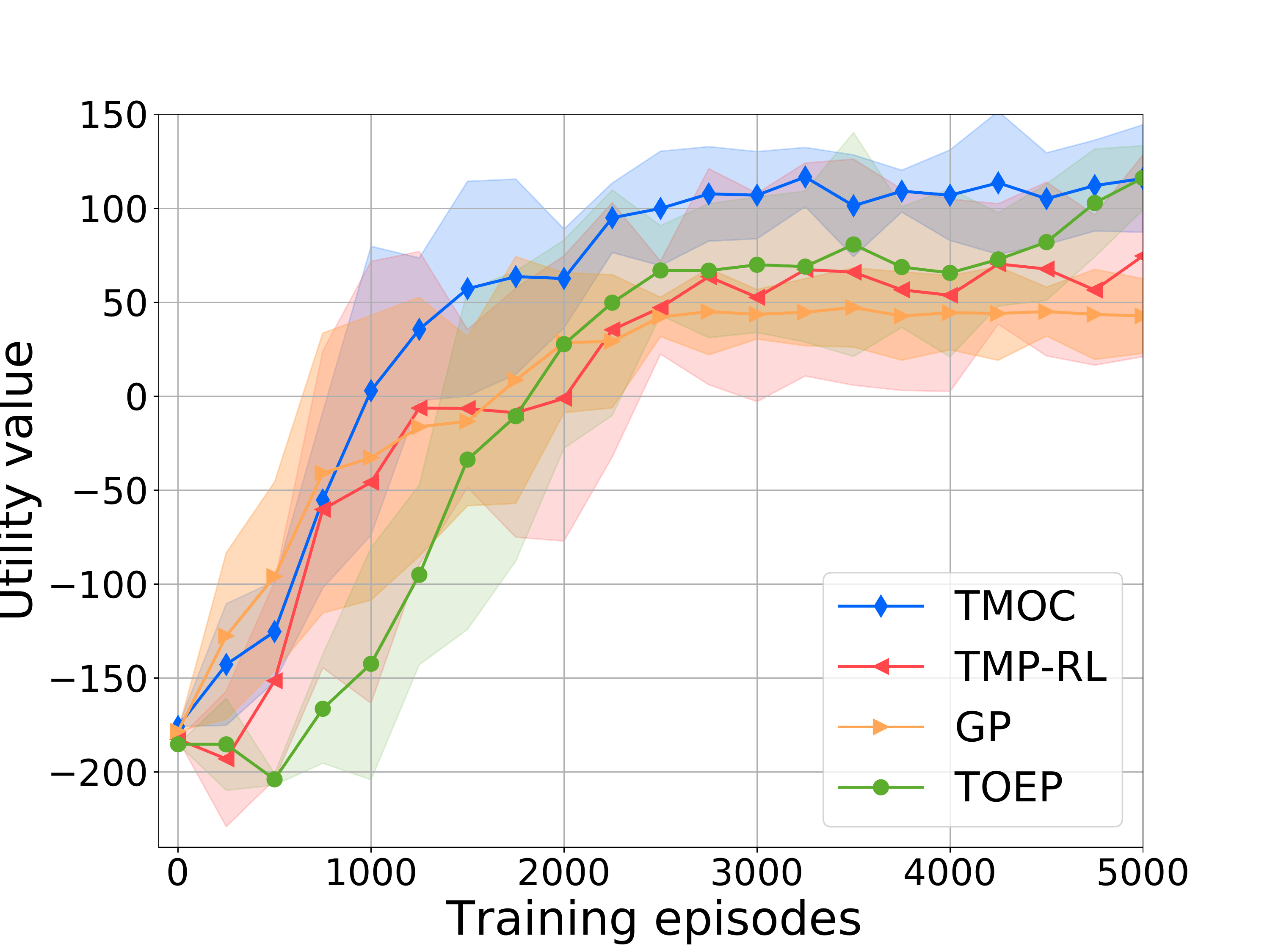}
\hspace{-1.2em}
\includegraphics[width=.69\columnwidth]{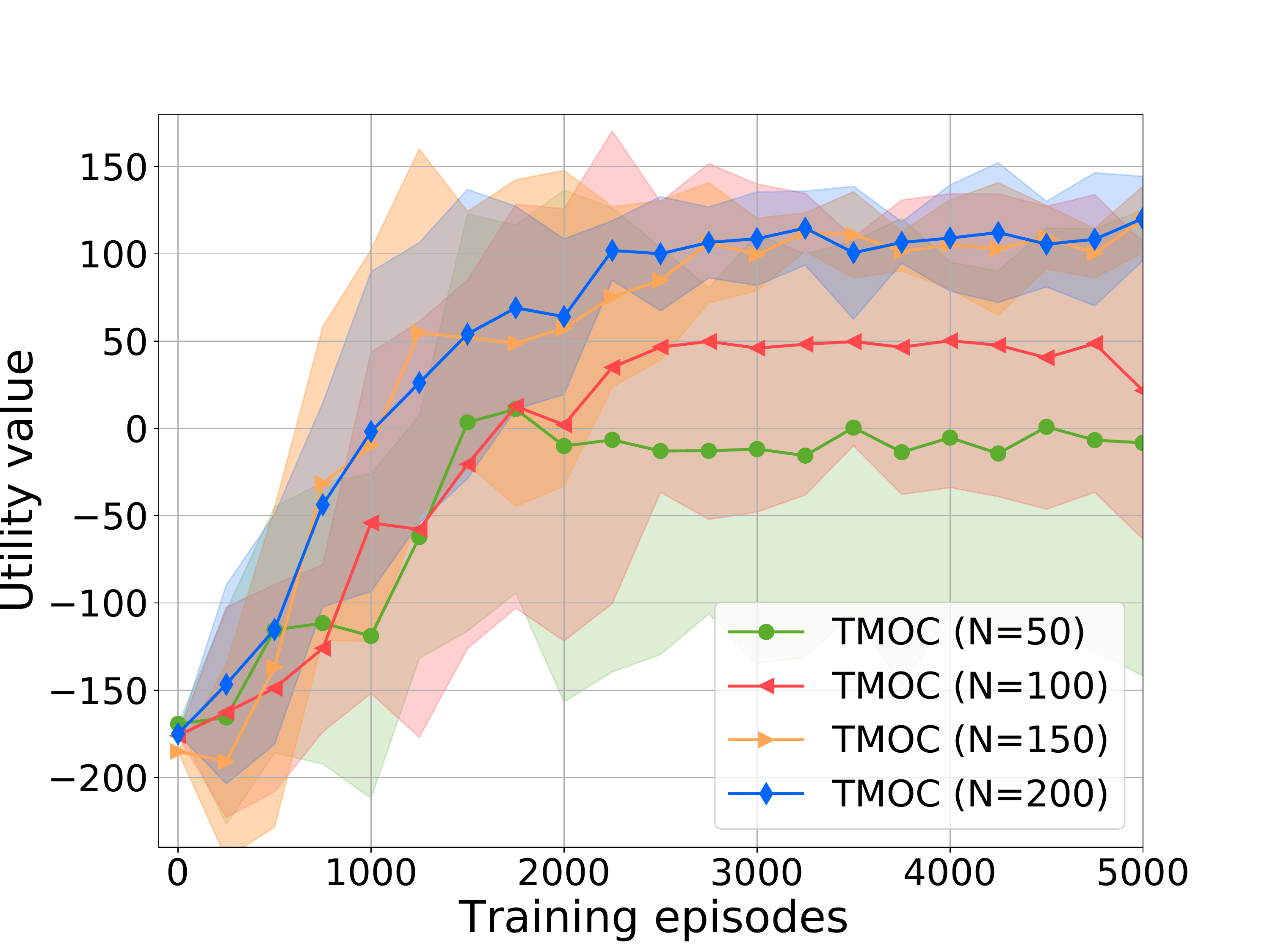}
\hspace{-1.2em}
\includegraphics[width=.69\columnwidth]{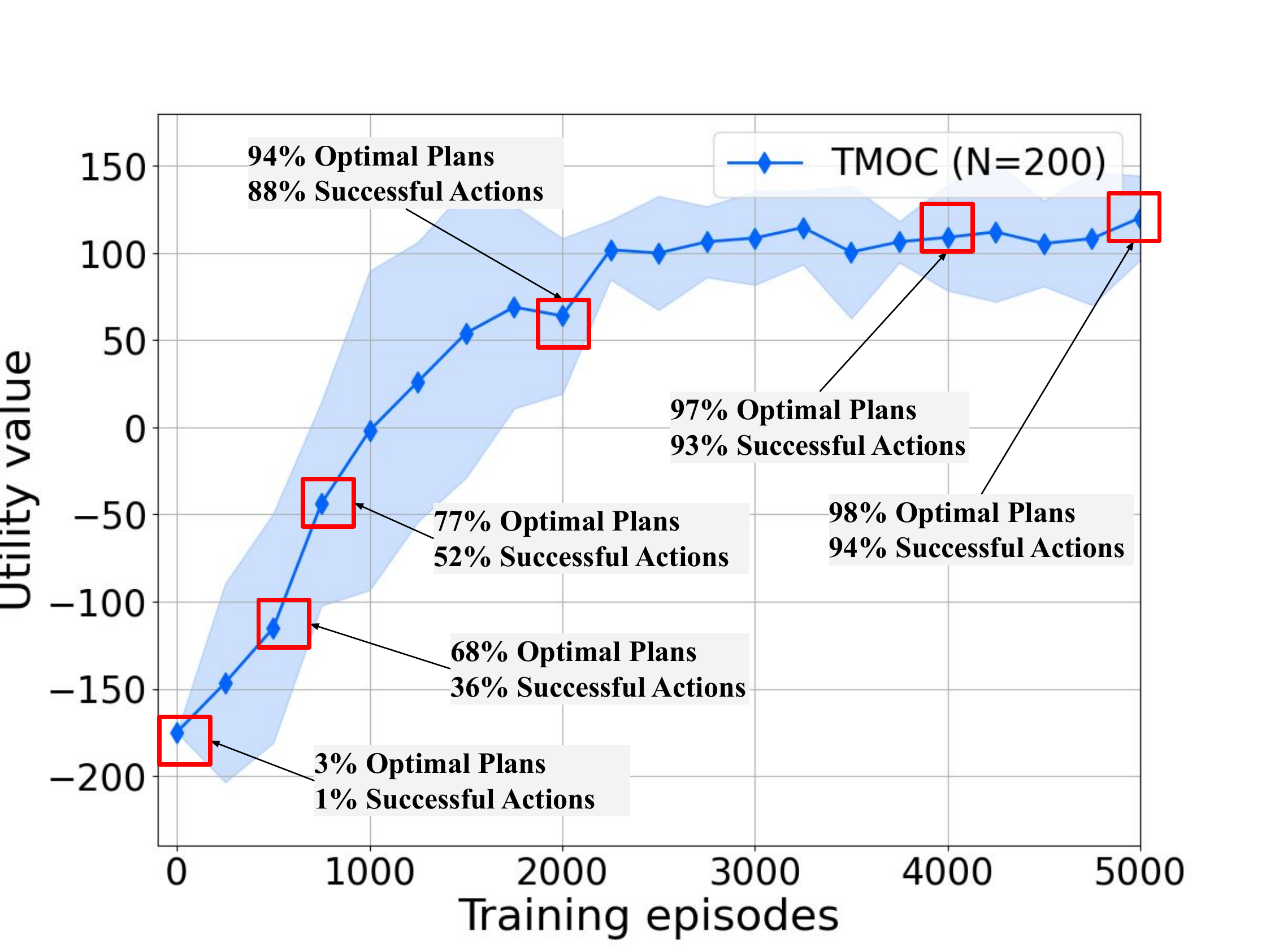}
\hspace{-2em}
\caption{\textbf{Left}: Comparisons between TMOC and three baselines of TMP-RL, GP and TOEP in the task of a gripper moving objects using a container; \textbf{Middle}: Performance of TMOC under different numbers of simulated worlds, where $N$ = 50, 100, 150, and 200; \textbf{Right}: Milestones of TMOC, where we present the percentages of optimal plans and successful actions. 
In all subfigures, the $y$-axis refers to the cumulative utility obtained by the real robot.}\label{fig:experiment_results}
\end{figure*}

\vspace{.5em}
\noindent
\textbf{Baselines:} 
Three baseline methods are utilized in this research, and they are selected from the literature, referred to as \textbf{TMP-RL}~\cite{jiang2019task},  \textbf{GP}~\cite{wang2021learning}, and \textbf{TOEP}~\cite{liang2020learning}.
\begin{itemize}
    \item TMP-RL is a task-motion planning algorithm that learns from trial and error. TMP-RL is not object-centric, and hence does not learn object properties ($L$) over time. 
    \item GP is a task-motion planning algorithm that learns primitive skills from trial and error, but cannot improve its task-level planning strategies. In our implementation of GP, the agent randomly selects one of the satisficing plans to achieve high-level goals. 
    \item TOEP does not learn the state mapping function, and the gripper is  placed (on a block) in a grasping position that is randomly selected in a reasonable range. It should be noted that our ``TOEP'' agent is equipped with task-planning capability, whereas the original TOEP work used predefined task-level behaviors. 
\end{itemize}

\vspace{.5em}
\noindent
\textbf{TMOC vs. Three Baselines:} 
For every approach, we conducted 15 runs with 5000
episodes in each run. 
Fig.~\ref{fig:experiment_results} (Left) illustrates the learning curves for utility value, averaged over the 15 runs with the shaded regions representing one standard deviation from the mean.
From the results, we get the important observation that TMOC \footnote{The number of grounded worlds in TMOC is 200 (i.e., $N=200$).} performs better than the baselines in terms of cumulative utility and learning rate. 

One interesting observation from Fig.~\ref{fig:experiment_results} (Left) is that TOEP fell behind the other three approaches (including our TMOC) at the beginning, and then later reached a cumulative utility level that is comparable to that of TMOC. 
This is because TOEP does not learn the state mapping function, and hence cannot improve its skills of ``grasping'' and ``pushing'' from trial and error. 
This disadvantage affects the learning rate of TOEP at the beginning. 
At the late learning phase (after about 2000 episodes), the TOEP agent was able to catch up, because the task planner learned to adapt to the motion planner (that is suboptimal on grasping and pushing). 
As a result, the ultimate performances of TOEP and TMOC (ours) are comparable. 
TMP-RL does not learn object properties ($L$), which affected its learning rate and cumulative utility level compared with  TMOC (ours), because an inaccurate $L$ is detrimental to the learning of both $Y$ and $T$.

\begin{figure*}[t]
\centering
\includegraphics[width=2.\columnwidth]{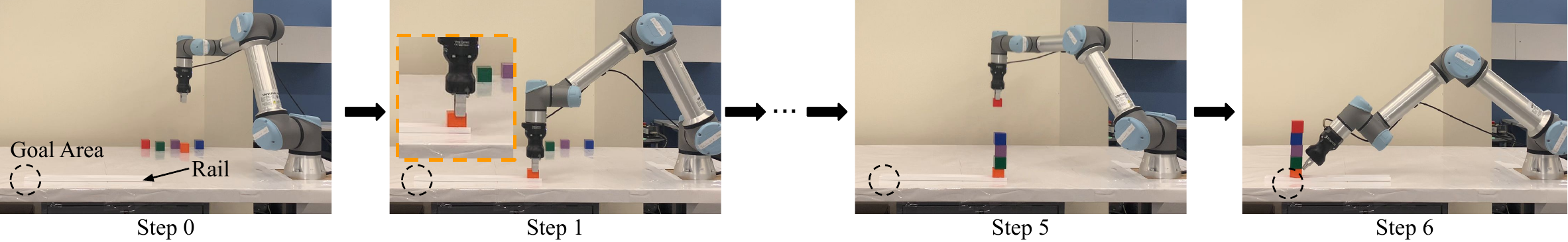}
\caption{A robot arm is tasked to move five blocks from an initial area. 
To perform ``2D'' experiments in the 3D real world, we attached a rail onto a table, where blocks can be only moved back and forth.
There is a trade-off between stability and efficiency (higher stacks are less stable) that the robot must learn from trial and error.
In this case, five blocks are stacked and then moved to the goal area. 
The robot was lucky enough to be successful in this trial, but might fail in other trials of performing the same plan. 
}\label{fig:demo_ur5e}
\vspace{-.5em}
\end{figure*}


\noindent
\textbf{Number of Grounded Worlds:} The performance of TMOC highly depends on the number of simulated worlds (or ``particles,'' in the terminology of Particle Filters), which is referred to in Line~\ref{alg:init_N} of Algorithm~\ref{alg:TMOC}. 
Each curve was based on 12 runs with 5000 episodes in each run.
Sufficient particles are necessary to ensure the quality of a Particle filter, and to represent the distributions being estimated.
We are interested in answering this question: \emph{How many particles are adequate for our experiments?}
Thus, we evaluated TMOC under different numbers of particles (i.e., $50 \leq N \leq 200$).

Fig.~\ref{fig:experiment_results} (Middle) shows no significant improvement in learning rate or cumulative utility when we increased the particle number from $N=150$ to $N=200$. 
Therefore, we believe $N=150$ is sufficient to the robot in our domain. 
This observation can serve as a reference to TMOC practitioners. 

\vspace{.5em}
\noindent
\textbf{Milestones of TMOC:} 
We are also interested in how the robot makes progress in learning $L$, $Y$, and $T$ while running TMOC. 
Thus, we calculated the percentages of optimal plans and successful actions at some selected training episodes, where $N=200$ in this experiment. 

Fig.~\ref{fig:experiment_results} (Right) shows that both percentages of optimal plans and successful actions are close to  0\%  at the beginning, because of the highly inaccurate functions of $L$, $Y$, and $T$.
They gradually increased and finally got close to 100\% after 5000 episodes.
At the same time, we see that $L$, $Y$, and $T$ have different learning rates. 
For instance, when the robot completed about 2000  episodes, $T$ converged well because most of the  plans (about 94\%) are optimal, while there is still room to further learn functions $L$ and $Y$ because only 88\% of actions are successful. 
We attribute the growth of utility value after 2000 episodes to the learning of $L$ and $Y$.

\vspace{.5em}
\noindent
\textbf{Task Variations of TMOC:}
The performance of TMOC is affected by task variations, such as blocks of different sizes. 
Thus, we changed the goal specification to let the robot stack and move one big block and seven small ones. 
The side length of the big one is 20\% bigger than the small ones.  
The robot knows that all blocks share the same density, but must estimate their sizes for task-motion planning. 
Introducing different sizes makes the stack-and-push task too difficult for the robot. For evaluation purposes, we provided the robot with guidance that the big block should be put near the bottom (in one of the first two steps). 

\begin{figure}[]
\vspace{-.5em}
\centering
\includegraphics[width=0.7\columnwidth]{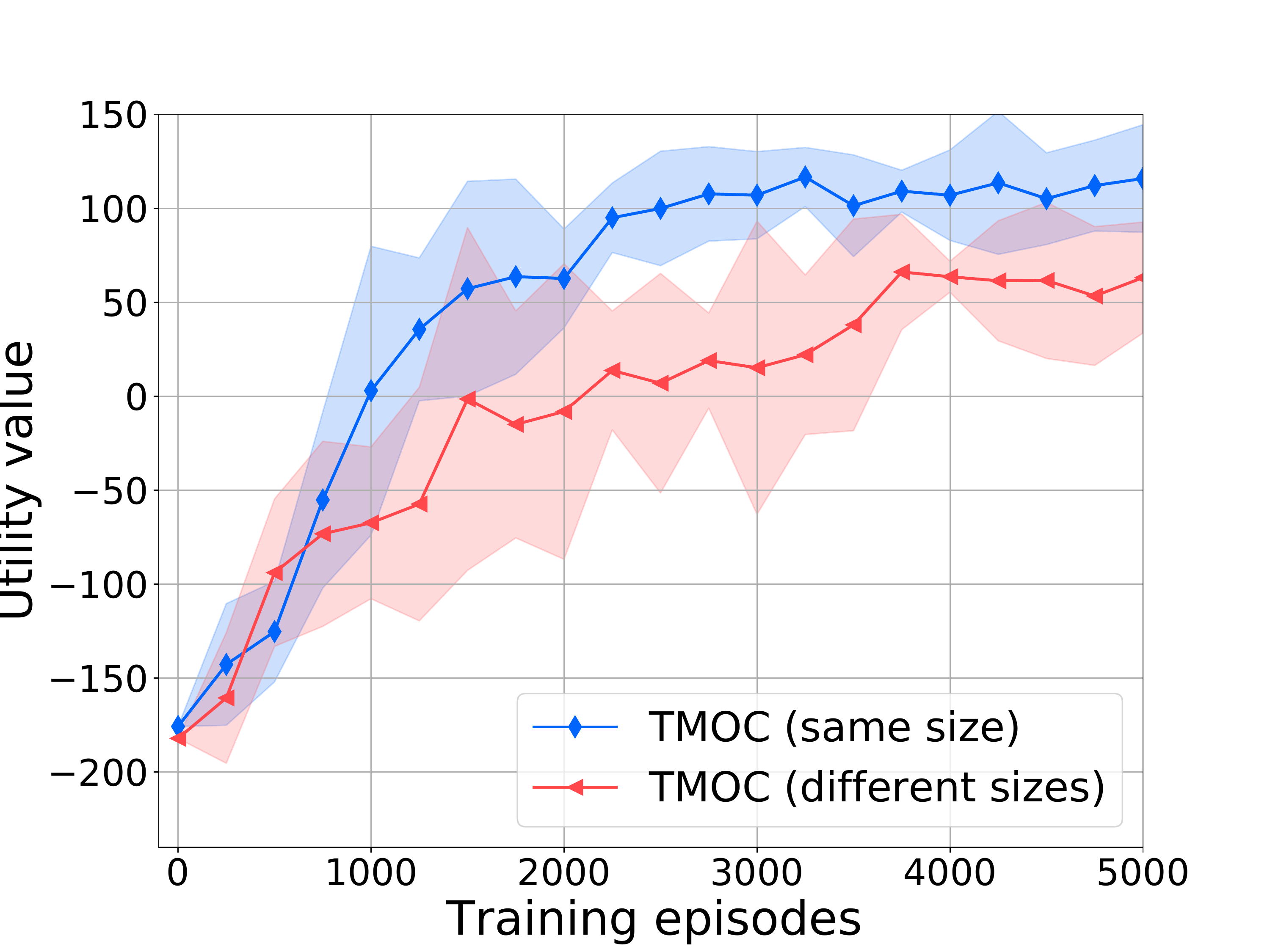}
\caption{Performance of TMOC under a task variation (i.e., size). \textbf{Same Size}: the robot is tasked to stack and move eight small blocks. \textbf{Different Sizes}: the robot is tasked to stack and move one big block and seven small ones. In both curves, $N$ equals $200$.}
\label{fig:size}
\end{figure}

Fig.~\ref{fig:size} shows that under the ``different sizes'' domain variation, TMOC needs more episodes to reach a utility level that was achieved when all blocks share the same shape. 
Also, under the ``different sizes'' variation, TMOC's convergence level was lower than that under the ``same size'' variation. 
We observed that grasps become more unreliable when the robot faces a bigger block.
Incidentally, we found no significant difference in the performance of TMOC with and without block weight variations.

\begin{table}[t]
\centering
\scriptsize
\caption{Utility value of TMOC under different episodes in the stack-and-push task using real robot arm UR5e}\label{tab:experiment_ur5e}
\begin{tabular}{c | c c c c c c c}
\bottomrule[1pt]
\textbf{Episode}& $0$ & $4$ & $8$ & $12$ & $16$\\ \hline 
Utility & $-23(18)$ & $60(128)$ &  $-10(32)$ &  $4(67)$ &  $183(80)$\\ \hline  \hline 
\textbf{Episode}    & $20$ & $24$ & $28$ & $32$ & $36$\\ \hline
Utility & $70(98)$ &  $86(113)$ &  $105(74)$ & $218(11)$ & $235(0)$ \\
\bottomrule[1pt]
\end{tabular}
\end{table}

\subsection{Experiments in the Real World}

TMOC was evaluated using a UR5e robotic arm.
In the experiment setting, the robot stacked blocks and then pushed them to a goal area, as illustrated in Fig.~\ref{fig:demo_ur5e}.
We provided the robot with accurate object properties (block size in this case) to control the difficulty of learning task-motion behaviors to a reasonable level. As a result, the robot only needed to learn a state mapping function (e.g., where to place the gripper for grasping) and a transition function (e.g., a stack-and-push strategy, including how many blocks should be stacked together).
We conducted three runs, each including 44 episodes, that together took more than 12 hours. 

Table~\ref{tab:experiment_ur5e} shows the mean utility values and the corresponding standard variances of the TMOC algorithm after different numbers of episodes.
At the beginning, the mean value is low while the variance is high, because the robot does not know how to grasp blocks or what a good action sequence is like. 
With more trials, our TMOC robot learned a good state mapping function $Y$ and transition function $T$, which resulted in a high utility value and a low variance.
More specifically, the robot learned reasonably good $Y$ and $T$ functions after 15 and 32 episodes, respectively.
The result demonstrates that learning $Y$ and $T$ enables the robot to complete the stack-and-push task efficiently and reliably, which is overall evaluated using the utility values.

\section{Conclusion and Future work}
In this paper, we develop a grounded task-motion planning algorithm TMOC that can ground objects (including their physical properties) for robot planning in mobile manipulation tasks that involve complex multi-object dynamic interactions (e.g., building blocks). 
We assume the availability of a high-fidelity simulator, and the unavailability of accurate properties of those objects (e.g., size, weight, and shape).
The robot needs to plan high-level behaviors to fulfill complex goals that require picking, moving, and placing the objects while minimizing overall action costs.
Thus, we develop the TMOC algorithm to learn the object properties, state mapping function and transition function.
Results from both the simulation and the real world demonstrate that TMOC improves the task-completion effectiveness and efficiency in terms of utility value and learning rate.

The paper assumes that all objects are relevant to the current task, and hence TMOC grounds all of them in physics-based simulation. 
In the future, we plan to enable the robot to learn what objects should be grounded and when. 
Objects have many different physical properties, and estimating their values requires different perception methods. 
This paper considered size and weight, and other properties can be estimated in future work. 
Another direction for future research is to further increase the number of particles (each corresponding to a grounded world) for more accurately estimating object properties, where if needed, distributed computational platforms can be used. 
We are also interested in applying TMOC to domains beyond ``stack and push'' in the future. 
Further, our real-robot experiment can be improved by including other everyday objects with diverse properties. 
Finally, it is important to look into the theoretical properties of TMOC, such as its completeness and scalability. 
Such theoretical analysis can be difficult due to the stochastic nature of TMOC (e.g., many grounded worlds) and the iterative learning process, and we leave it to future work.




\bibliographystyle{IEEEtran}
\bibliography{ref}
\end{document}